\documentclass[10pt,twocolumn,letterpaper]{article}

\usepackage{cvpr}              

\usepackage[dvipsnames]{xcolor}

\definecolor{cvprblue}{rgb}{0.21,0.49,0.74}
\usepackage[pagebackref,breaklinks,colorlinks,citecolor=cvprblue]{hyperref}

\def\paperID{*****} 
\def\confName{CVPR}
\def\confYear{2024}

\usepackage{ulem}
\usepackage{tcolorbox}
\usepackage{lipsum}
\usepackage{bbding}
\usepackage{url}
\usepackage{adjustbox} 

\usepackage{color}
\usepackage{enumitem}

\usepackage{multirow}
\usepackage{ulem}

\usepackage{url}            
\usepackage{tcolorbox}       
\usepackage{amsfonts}       
\usepackage{nicefrac}       
\usepackage{microtype}      
\usepackage{xcolor}

\usepackage{float}
\definecolor{myred}{RGB}{214,45,44}
\definecolor{myblue}{RGB}{26,118,181}
\definecolor{mygreen}{RGB}{63,159,52}
\definecolor{myorange}{RGB}{251,125,14}
\usepackage[section]{placeins}
\usepackage{hyperref}
\usepackage{url}
\usepackage{graphicx}

\usepackage{booktabs}
\usepackage{tablefootnote}
\usepackage[flushleft]{threeparttable}

\usepackage{amsmath}
\usepackage{amsfonts}
\usepackage{amssymb}
\usepackage{wrapfig}
\usepackage{subcaption}
\usepackage{multirow}
 \usepackage{mathtools} 

\usepackage{verbatim}

\usepackage{anyfontsize}

\usepackage{microtype}
\usepackage{graphicx}
\usepackage{booktabs} 
\usepackage{multirow}
\usepackage{amsmath,amssymb}
\usepackage{booktabs}
\usepackage{caption,subcaption}

\usepackage{xcolor}
\definecolor{mygreen}{HTML}{3cb44b}
\definecolor{skyblue}{HTML}{beffff}
\definecolor{lightgreen}{HTML}{90ee90}

\usepackage{color, colortbl}

\definecolor{emerald}{rgb}{0.31, 0.78, 0.37}

\usepackage{tcolorbox}
\usepackage{enumitem}
\setitemize{itemsep=10pt,topsep=0pt,parsep=0pt,partopsep=0pt}
\usepackage{colortbl}

\usepackage{xcolor}
\definecolor{mygreen}{HTML}{3cb44b}
\colorlet{myyellow}{green!10!orange!90!}
\makeatletter

\usepackage{tikz}
\usetikzlibrary{arrows,shapes,snakes,automata,backgrounds,fit,petri}
\usepackage{adjustbox}

\newcommand{\RN}[1]{%
	\textup{\lowercase\expandafter{\it \romannumeral#1}}%
}
\usepackage{tabu}

\newcommand{\beq}{\vspace{0mm}\begin{equation}}
\newcommand{\eeq}{\vspace{0mm}\end{equation}}
\newcommand{\beqs}{\vspace{0mm}\begin{eqnarray}}
\newcommand{\eeqs}{\vspace{0mm}\end{eqnarray}}
\newcommand{\barr}{\begin{array}}
\newcommand{\earr}{\end{array}}

\usepackage{color, colortbl}
\definecolor{Gray}{gray}{0.93}

\usepackage{lipsum}

\usepackage{pifont}

\usepackage{makecell}

\usepackage{xcolor,amsmath}
\usepackage[linesnumbered,ruled,vlined]{algorithm2e}
\DontPrintSemicolon

\usepackage{xcolor}
\definecolor{mygreen}{HTML}{3cb44b}

\SetKwComment{Comment}{\color{green!50!black}\# }{}

\newcommand{\var}{\texttt}

\SetKwProg{Function}{def}{:}{}

\SetKwProg{For}{for}{:}{}
\SetKwProg{If}{if}{:}{}
\newcommand{\VarSty}[1]{\textnormal{\ttfamily\color{blue!90!black}#1}\unskip}

\newcommand{\mypara}[1]{\vspace{0.2cm}\noindent\textbf{#1}\hspace{0.1cm}}

\title{LISA++: An Improved Baseline for Reasoning Segmentation \\with Large Language Model}

\author{
Senqiao Yang$^{1*}$
~
Tianyuan Qu$^{1*}$
~
Xin Lai$^{1*}$
~
Zhuotao Tian$^{2\dagger}$
~
Bohao Peng$^{1}$
~
Shu Liu$^{2}$
~
Jiaya Jia$^{1,2}$
\\[0.2cm]
\small{
$^1$The Chinese University of Hong Kong~~
$^2$SmartMore~~ 
}
}

\begin{document}
\maketitle
\footnotetext[1]{~*~Equal contribution.}
\footnotetext[2]{~$\dagger$~Correspondence to \url{tianzhuotao@gmail.com}.}
\footnotetext[3]{~Work in progress. As the supplementary to show LISA~\cite{lai2023lisa} can be decently extended to more complex scenarios without any structural change.}
\begin{abstract}

While LISA effectively bridges the gap between segmentation and large language models to enable reasoning segmentation, it poses certain limitations: unable to distinguish different instances of the target region, and constrained by the pre-defined textual response formats. In this work, we introduce LISA++, an update to the existing LISA model, focusing on improving core functionalities while keeping the base architecture intact. The main enhancements in LISA++ include:
\textbf{1) Enhanced Segmentation}: The instance segmentation ability has been added, providing a more detailed scene analysis along with the existing multi-region semantic segmentation.
\textbf{2) More Natural Conversation}: Improved capability for multi-turn dialogue, with the ability to incorporate segmentation results directly into text responses, i.e., Segmentation in Dialogue (SiD).
These improvements are achieved by curating the existing samples of generic segmentation datasets, aimed specifically at enhancing the segmentation and conversational skills without structural change and additional data sources.
Comparative analysis with the original LISA model shows significant advancements in these areas, positioning LISA++ as a notable upgrade in visual understanding and interaction. LISA++'s adaptability and improved features highlight the versatility of the mask-as-embedding paradigm proposed by LISA, and the potential as a foundational model for diverse applications.
\end{abstract}    
\section{Introduction}
\label{sec:intro}

\begin{figure}[th!]
	\centering
        \includegraphics [width=1\linewidth] 
        {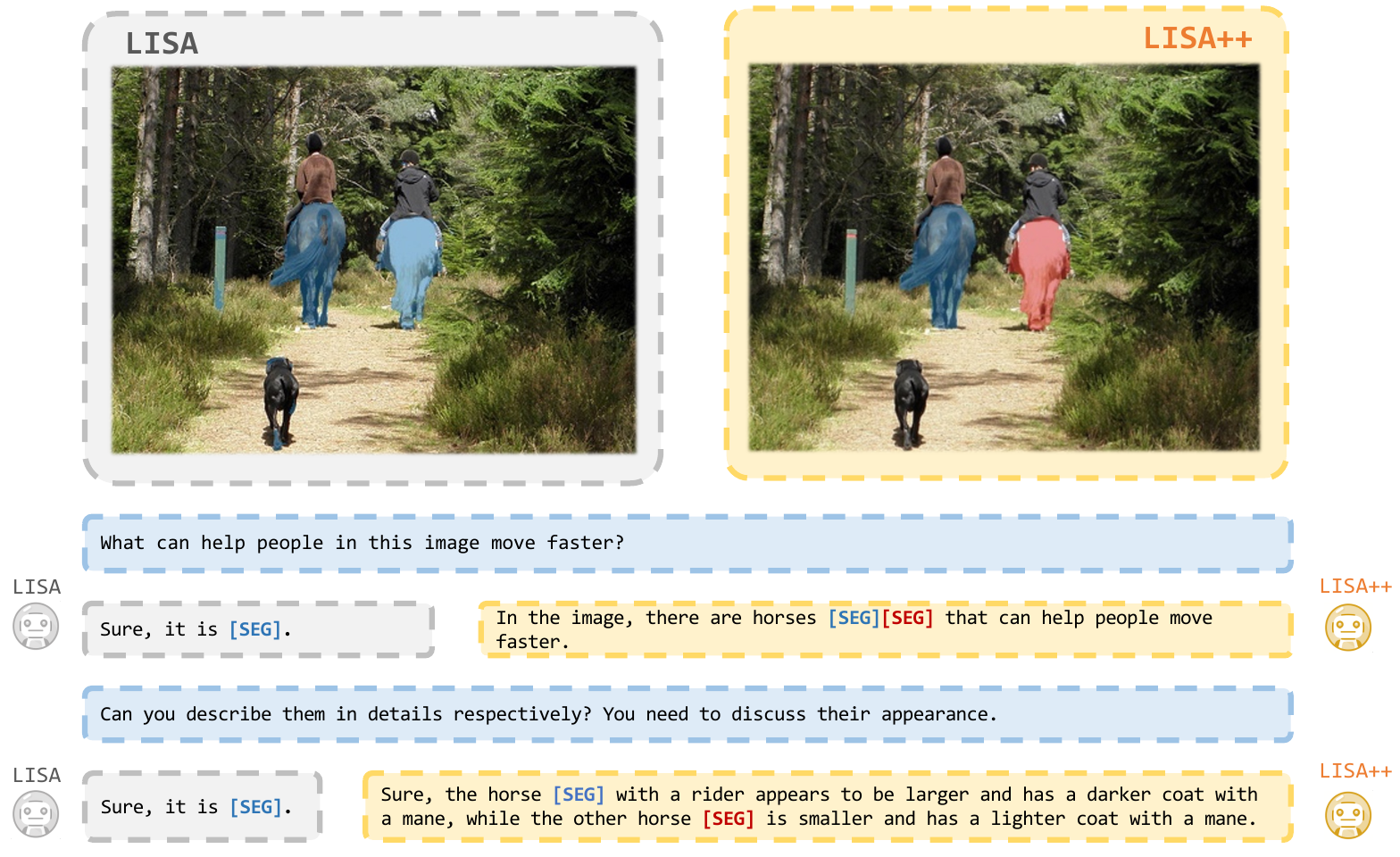}
    
    \caption{A multi-turn conversation of LISA (left) and LISA++ (right). LISA++ demonstrates enhanced capabilities compared to LISA. Particularly, LISA++ can differentiate different individuals within the same category. Also, LISA++ exhibits superior segmentation proficiency in the dialogue, resulting in more natural and flexible conversations. Both models are asked to segment with explanations, and the instruction templates are omitted.}
    \label{fig:teaser}
\end{figure}

Visual perception tasks~\citep{zhang2022deep,luo2023pfenet++,tian2022adaptive} function as essential tools enabling machines to comprehend and interpret their environment. Yet, the single-source visual data restricts their further applications where information derived from additional modalities should be considered.

Multimodal models~\citep{koh2023grounding,wu2023visual,yang2023mm,shen2023hugginggpt,liu2023internchat,yang2023gpt4tools,wu2023ppt,li2023blip,ye2023mplug,wu2023see,li2023otter} have increasingly become pivotal in the recent surge of advancements in artificial intelligence, serving as foundational elements in the development of versatile general-purpose assistants. As these large multimodal models~\cite{llama,zheng2023judging} continue to evolve, their contribution becomes increasingly central in expanding the horizons of AI research and capabilities. Exemplars such as LLaVA~\cite{liu2023visual,liu2023improved}, MiniGPT~\cite{zhu2023minigpt,chen2023minigptv2} have demonstrated exceptional proficiencies in adhering to natural instructions and visual reasoning.

Nevertheless, these widely recognized multimodal models predominantly generate textual responses based on images and queries, lacking the capacity to provide detailed positional information, \textit{e.g.}, bounding boxes and segmentation masks, about specific entities within images. This limitation curtails their applicability in more sophisticated tasks. In response to this gap, LISA~\citep{lai2023lisa} was conceived as a means to bridge large language models (LLMs) with segmentation tasks. LISA proposes the novel \textit{mask-as-embedding} paradigm, enabling it to output segmentation masks of the areas of interest in images in response to both straightforward and intricate user queries. Concurrently, it can also provide comprehensible textual explanations, thereby marking a significant stride towards leveraging LLMs in downstream applications with dense visual perception.

In parallel endeavors, some contemporaneous works~\citep{griffon2023zhan,lenna2023wei,peng2023kosmos,detgpt,perceptiongpt} with LISA integrate object detection with extensive models to enhance perception capabilities. Nonetheless, LISA distinguishes itself by generating detailed segmentation masks, thereby offering a more intricate understanding of images compared to the traditional bounding boxes generated by detection frameworks. However, the existing version of LISA presents certain limitations: 1) Though adept at generating single or multi-region segmentation results, it lacks proficiency in instance segmentation, notably in distinguishing between objects of the same category; 2) Moreover, while capable of producing a segmentation mask and explanation based on the input query and image, LISA's output format is relatively rigid and uniform. Typically, LISA's responses commence with a generic ``\texttt{\small Sure, it is \texttt{<SEG>}}'', followed by an analysis or explanation of the object linked to the \texttt{<SEG>} token, detracting from the naturalness and adaptability required by the real applications.

To rectify these issues, we introduce LISA++ in this paper, an advancement of LISA. Notably, LISA++ retains the original architectural framework of LISA. By only reconstructing the instruction-tuning data using prevalent segmentation datasets such as COCO~\citep{caesar2018coco} and ADE20K~\citep{zhou2017scene}, LISA++ enables LISA to support both semantic and instance segmentation tasks. Furthermore, LISA++ can yield more natural and flexible responses, by integrating segmentation results into textual responses, termed the ability of Segmentation in Dialogue (SiD). 
An example of a multi-turn conversation is shown in Figure~\ref{fig:teaser} which highlights the limitations of LISA in this scenario, whereas LISA++ demonstrates its capability to identify different individuals with the same category and perform segmentation within the dialogue. 

Besides, we expand the original ReasonSeg dataset, which evaluates only a single target for each query, by including an additional benchmark for multi-target instance segmentation. This refinement allows for comprehensive assessments spanning both semantic and instance segmentation. 
As the \textit{mask-as-embedding} paradigm proposed by LISA~\citep{lai2023lisa} has shown to be simple, general, and effective, we hope that this paradigm, along with LISA++ and the updated ReasonSeg benchmark, will inspire future work to explore the potential of leveraging it to empower large models with even more capabilities.

\section{Background}
\label{sec:background}

\mypara{Large Vision-language Models (VLMs). }
Driven by the impressive reasoning capabilities exhibited by Large Language Models (LLMs), researchers are actively exploring methods to extend these proficiencies into the realm of visual perception, giving rise to the development of multimodal LLMs. An exemplary instance, Flamingo~\citep{alayrac2022flamingo}, employs a cross-attention architecture to focus on visual contexts, thereby facilitating context-aware visual learning. LLaVA~\citep{liu2023visual,liu2023improved}, MiniGPT-4~\citep{zhu2023minigpt} align the image and text features and then proceed with instruction tuning. Other models like BLIP-2~\citep{li2023blip} and mPLUG-OWL~\citep{ye2023mplug} put forth strategies for encoding image features through a visual encoder, which are then integrated with textual embeddings within the LLM framework. However, these VLMs, even the cutting-edge GPT-4V~\citep{gpt-4v}, can only provide text outputs, inhibiting the complex applications where dense visual perception is needed.

\mypara{VLMs with Dense Perception. }
In recent investigations, there has been a growing focus on the convergence of foundation models and the tasks related to dense visual perception. For example,~\citep{li2023semantic,zou2023segment,koh2023grounding} leverage the CLIP pre-trained foundation models to enable open-world segmentation and detection, but they are unable to handle complex instructions. Differently, VisionLLM~\citep{wang2023visionllm} combines a range of vision-centric tasks by utilizing instruction tuning with Large Language Models (LLMs). However, it may fall short of fully harnessing the potential of LLMs for handling intricate reasoning tasks.
In parallel research efforts, grounding capabilities and open-vocabularies detectors are leveraged by Kosmos-2~\citep{peng2023kosmos}, Qwen-VL~\citep{Qwen-VL} and DetGPT~\citep{detgpt}, enabling user-guided detection.
Moreover, GPT4RoI~\citep{zhang2023gpt4roi}, Ferret~\citep{you2023ferret}, and Shikra~\citep{chen2023shikra} introduce spatial boxes as input and train the model using region-text pairs, offering regional image understanding. LISA~\citep{lai2023lisa}, however, takes a different approach by focusing on the efficient integration of segmentation capabilities into VLMs through the novel \textit{mask-as-embedding paradigm} that enables the end-to-end training pipeline. 
\section{Method}
\label{sec:method}

In the following, Section~\ref{sec:newmodel} presents the details of the proposed LISA++ that additionally incorporates instance segmentation ability and the more natural and flexible segmentation in dialogue ability to the original LISA. More details of LISA and the setting of reasoning segmentation can be found in~\citep{lai2023lisa}. 
Then, Section~\ref{sec:newdata} supplements the ReasonSeg benchmark with the new evaluation on the instance level.

\begin{figure*}[th!]
	\centering
    \begin{minipage}   {0.85\linewidth}
        \centering
        \includegraphics [width=1\linewidth] 
        {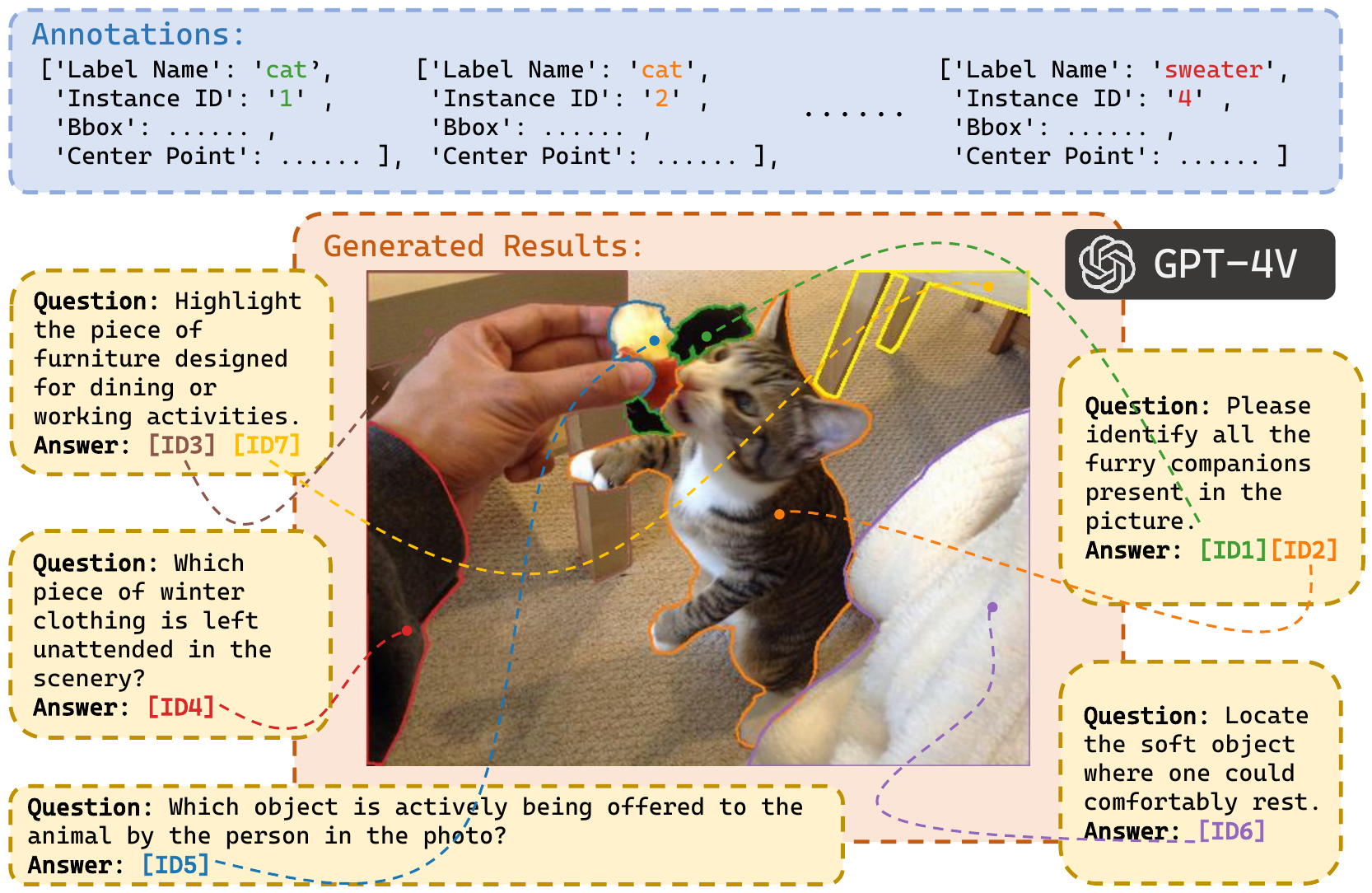}
    \end{minipage}    
    
    
    \caption{Illustration of Reasoning Instance Segmentation. Instance annotations and figures are sent to GPT-4V along with the carefully designed prompts, such that questions and answers regarding interesting instances can be yielded by GPT-4V automatically. The detailed prompt used is shown in Table~\ref{tab:prompt_reasonseg_inst} at the end of this manuscript.}
    \label{fig:reason_inst}
\end{figure*}

\subsection{An Improved Baseline}
\label{sec:newmodel}
This section introduces LISA++ which updates LISA with 1) enhanced reasoning instance segmentation ability and 2) more flexible and natural text responses with the ability of Segmentation in Dialogue (SiD). These two features are achieved by constructing the corresponding instruction tuning pairs. Details are as follows.

\mypara{Reasoning Instance Segmentation. }
LISA cannot perform instance segmentation (InstSeg) during the reasoning process, due to its limited ability to differentiate individuals within the same category. Recognizing that instance segmentation offers a more detailed visual perception capability, it is essential to enhance the model to adeptly perform both semantic and instance segmentation. This adaptation should be flexible to align with specific user instructions, enabling a more nuanced and responsive visual analysis.

To accomplish this objective, we utilize datasets with instance segmentation annotations as the foundation for instruction tuning. Specifically, these annotations are incorporated into prompts for GPT-4V, enabling it to generate insightful questions that reflect the detailed information in the annotations. 

For instance, consider the scenario shown in Figure~\ref{fig:reason_inst}: 
\textit{``two kittens, one white and one black, are standing up to eat the food their owner is feeding them''}. The image and the instance annotations are sent to GPT-4V, and for these cats, GPT-4V generates: ``\texttt{\textbf{Question}}: \texttt{\small Please identify all the furry companions present in this picture.} \texttt{\textbf{Answer}}: \textcolor{mygreen}{[\textbf{ID1}]} \textcolor{myorange}{[\textbf{ID2}]}" where \textcolor{mygreen}{[\textbf{ID1}]}  and \textcolor{myorange}{[\textbf{ID2}]} represent the mask annotations of two cats in the image, respectively. Besides, other surroundings are also queried, such as the \textcolor{myblue}{\textbf{food}} \textcolor{myblue}{[\textbf{ID5}]} (\texttt{\small Which object is actively being offered to the animal by the person in the photo?}) and \textcolor{myred}{\textbf{sweater}}  \textcolor{myred}{[\textbf{ID4}]} (\texttt{\small which piece of winter clothing is left unattended in the scenery?}).

\begin{figure*}[t!]
	\centering

            \includegraphics [width=0.8\linewidth] 
        {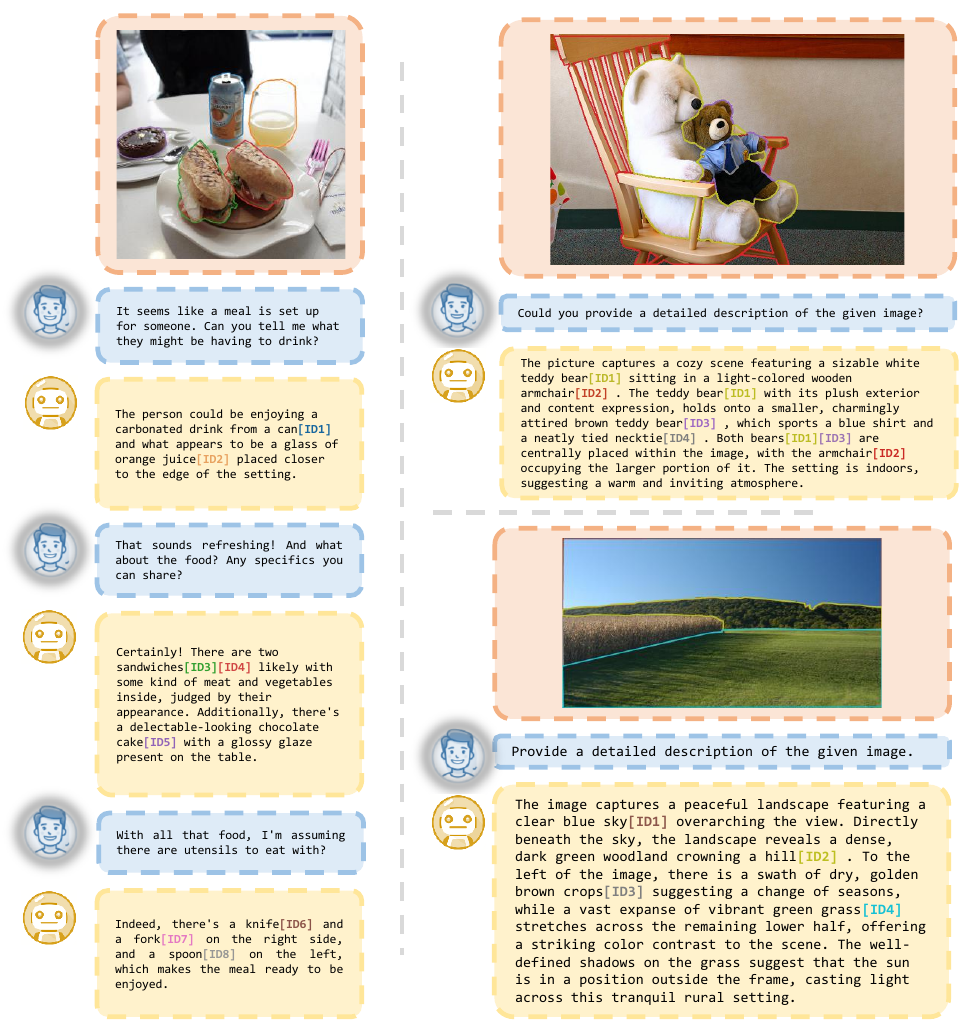}
    
    \caption{Illustration of two scenarios of Segmentation in Dialogue. LISA++ is required to perform natural conversations while yielding masks for different things and stuff simultaneously.}
    \label{fig:sid}
\end{figure*}

\mypara{Segmentation in Dialogue (SiD). }
As previously highlighted, LISA's ability to engage in natural and fluid conversations with users is hindered by its reliance on pre-defined response formats. To address this limitation, we leverage the segmentation annotations (whether they are instance segmentations or not) to facilitate the seamless integration of segmentation results, indicated as \texttt{<SEG>}, into the dialogue at contextually appropriate points, rather than confining them to a predetermined, static position in the response. We refer to this new capability as the Segmentation in Dialogue (SiD) that ensures natural and dynamic interactions between the AI agent and its users, enhancing the overall responsiveness and relevance of the dialogue. 
Concretely, we consider two scenarios regarding SiD: 1) reasoning Q\&A and 2) captioning, and the examples are shown in Figure~\ref{fig:sid}. The prompts are shown in Tables~\ref{tab:prompt_qa} and~\ref{tab:prompt_caption}.

\mypara{Integration of SemSeg, InstSeg and SiD.}
As instance segmentation is performed by LISA++, similar to the multi-region responses, multiple \texttt{<SEG>} tokens are yielded. However, different from LISA where different \texttt{<SEG>} tokens represent different categories, in LISA++, they may also represent different individuals within the same category. Therefore, during training, we adopt the \textit{bipartite matching}, as done in MaskFormer and DETR~\citep{cheng2022masked,carion2020end}, to find the ground truth for each predicted instance mask.

Nevertheless, an additional challenge arises in ensuring that LISA++ is compatible with both semantic segmentation (SemSeg) and instance segmentation (InstSeg), while seamlessly integrating the outcomes of these segmentations into the text responses generated through Segmentation in Dialogue (SiD). To achieve a cohesive amalgamation of these elements, we accordingly employ the task templates appended to user instructions. These templates are designed to facilitate the harmonious integration of two types of segmentation results into the dialogue framework, thereby enhancing the model's ability to comprehend and respond to complex segmentation tasks within a natural conversational context. The task templates are shown in Table~\ref{tab:text_template}.

\begin{table}[t!]

\centering
\begin{minipage}{0.99\columnwidth}\vspace{0mm}    \centering
\begin{tcolorbox} 
    \centering
    \small
     \hspace{-6mm}
    \begin{tabular}{p{0.99\columnwidth}}

\begin{minipage}{0.99\columnwidth}\vspace{0mm}
\VarSty{SemSeg} = ``The mask(s) are for semantic segmentation. No need to differentiate different instances within the same category."\\

\VarSty{InstSeg} = ``The mask(s) are for instance segmentation. Different instances within the same countable category should be predicted by separated masks. Uncountable category does not need separate masks."\\

\VarSty{SiD with SemSeg} = ``Please answer the question with text and output semantic segmentation mask prediction(s). No need to differentiate different instances within the same category."\\

\VarSty{SiD with InstSeg} = ``Please answer the question with text and output the instance segmentation mask prediction(s). Different instances within the same countable category should be predicted by separated masks. Uncountable category does not need separate masks."\\

\VarSty{Pure Conversation} = ``Please answer the question only with text, do not output mask."
\end{minipage}
    \end{tabular}
\end{tcolorbox}

\vspace{-2mm}
\end{minipage}
\caption{The instruction templates used for different tasks. They are appended to the user's instructions as part of the prompts. }
\label{tab:text_template}
\end{table}

As aforementioned, the captioning and Q\&A data used for SiD are annotated for instance segmentation in dialogue (SiD + InstSeg). Yet, it is important to note that they can be easily transformed into formats suitable for ``SiD + SemSeg" or purely textual dialogues. To be more specific, for ``SiD + SemSeg", one may only need to merge the instance annotations within the same category to one single annotation. On the other hand, to adapt the SiD data for pure textual conversation, one may simply remove all \texttt{<SEG>} tokens from the ground truth. By adopting these transformations during training, LISA++ is adept at encompassing a broader spectrum of potential conversational contexts, thereby enhancing its applicability and versatility in diverse interaction scenarios. To this end, LISA++ is capable of both semantic and instance segmentation and can seamlessly embed segmentation results into the responses in a controllable manner.

\subsection{Instance-level Evaluation for ReasonSeg}
\label{sec:newdata}
The ReasonSeg~\citep{lai2023lisa} benchmark evaluates the capability to pinpoint the target region in an image, responding to a complex text query. However, it is constrained as it considers only a single target region for each image-query pair.

To more thoroughly assess the model's reasoning segmentation capabilities, we introduce ReasonSeg-Inst, an extension of the ReasonSeg benchmark. This subset is specifically designed to evaluate the model's proficiency in instance discrimination. The previous version of ReasonSeg is named ReasonSeg-Sem because it only considers semantic segmentation and does not require differentiating between different instances within the same category. For more details of ReasonSeg-Sem, please refer to~\citep{lai2023lisa}.

\begin{figure*}[t!]
	\centering
    \begin{minipage}   {0.8\linewidth}
        \centering
        \includegraphics [width=1\linewidth] 
        {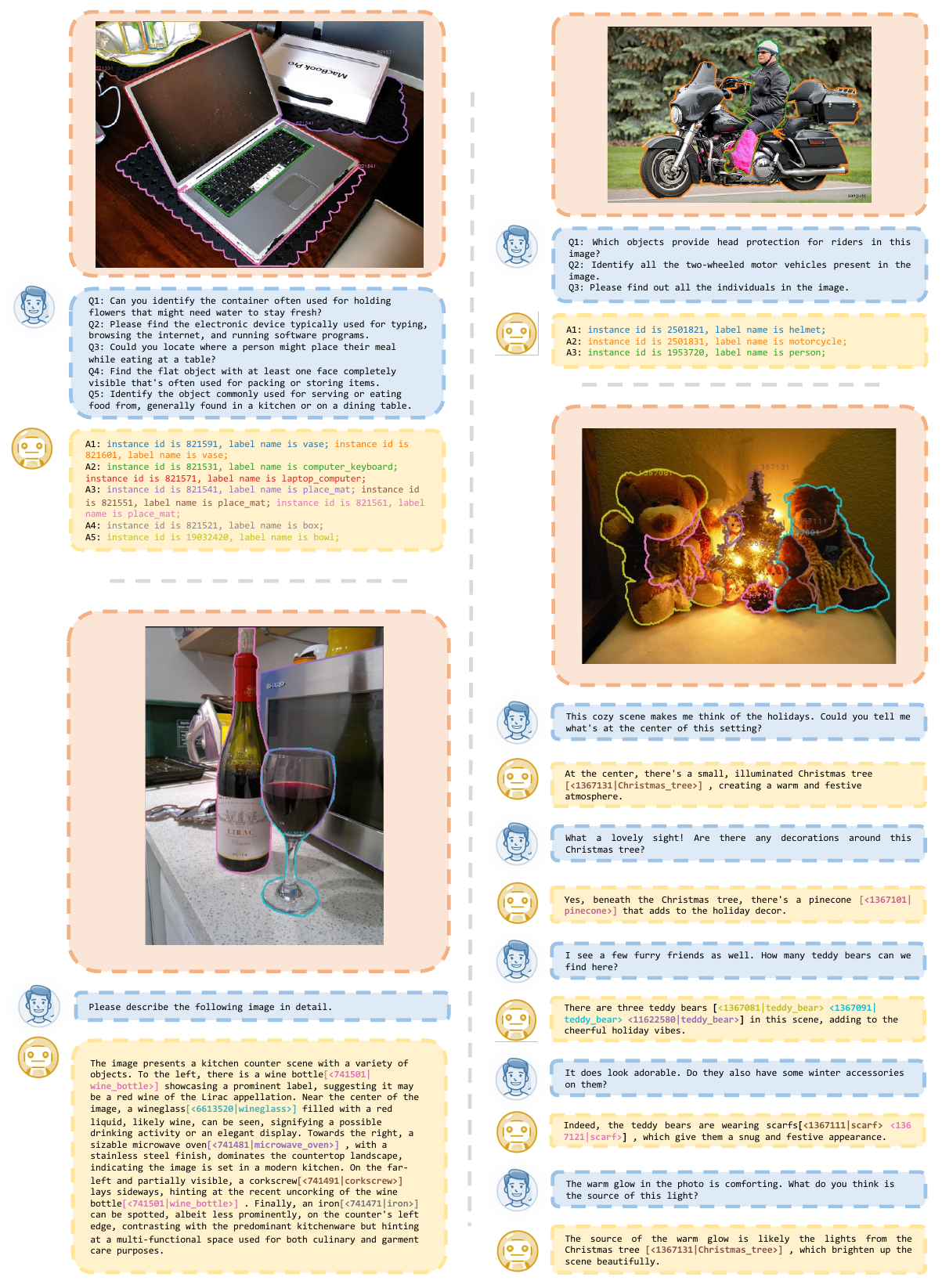}
    \end{minipage}    
    
    
    \caption{More examples of the reasoning instance segmentation and segmentation in dialogue.}
    \label{fig:more_examples}
\end{figure*}

The ReasonSeg-Inst dataset is curated from samples of COCO~\cite{caesar2018coco} and ADE20K~\cite{zhou2017scene} databases. Based on the aforementioned process for prompting GPT-4V, finally, we have collected 62K image-query pairs for the training phase, and 1.8K for evaluation. It has come to our attention that the low-resolution images or tiny object instances within COCO and ADE20K can impede the GPT-4V's ability to generate satisfactory instruction-tuning data. In light of this, we have adopted a filtration criterion, discarding the images whose size is below 512x512 and objects whose area is under 400 square pixels, to ensure the integrity of the dataset for effective instruction tuning data generation. More examples are shown in Figure~\ref{fig:more_examples} at the end of this manuscript.

\section{Experiments}
\label{sec:experiment}

\subsection{Implementation Details}
\mypara{Training. }
The training configuration for LISA++, such as training epochs, learning rate, losses and loss weights, etc., is predominantly in line with the framework of the original LISA model~\cite{lai2023lisa}. The notable divergence in LISA++ is the incorporation of new instruction tuning sets, specifically designed for refining reasoning in instance segmentation and SiD. To accomplish this, we utilize the COCO~\cite{caesar2018coco} datasets. It is important to note that the images from COCO were already employed in the training of the original LISA model, hence no additional visual data is introduced for training LISA++.

\mypara{Evaluation. }
For evaluation on ReasonSeg-Sem, we follow the setting of the original LISA~\citep{lai2023lisa} where cIoU and gIoU are reported. 
As for ReasonSeg-Inst which requires instance segmentation, we report the standard COCO metrics including AP, AP50, AP75, AP-small, AP-medium and AP-large for the evaluation of ReasonSeg-Inst. AP is evaluating using mask IoU.

\begin{table}[t!]
    \centering
    \tabcolsep=0.2cm
    {
    \begin{adjustbox}{width=1\linewidth,center=\linewidth}

        \begin{tabular}{ l  l | c c c  }
            \toprule
            
            
             &\multirow{2}*{Metric} & \multicolumn{3}{c}{Method}  \\

            \specialrule{0em}{0pt}{1pt}
            \cline{3-5}
            \specialrule{0em}{0pt}{1pt}            
            & & LISA-7B & LISA-13B & LISA++-7B  \\ 
            
            \specialrule{0em}{0pt}{1pt}
            \cline{1-5}
            \specialrule{0em}{0pt}{1pt}            

            
             & AP50 & 13.7 & 15.7 & 34.1  \\
             & AP75 & 6.6 & 8.1 & 22.1   \\
             & mAP  & 7.2 & 8.6 & 21.5  \\
             & AP-small & 0.9 & 1.0 & 2.5  \\
             & AP-medium &3.3 & 5.0 & 16.7\\
             & AP-large  & 19.4 & 21.3 & 47.1  \\
            \bottomrule            
        \end{tabular}
        \end{adjustbox}

    }
    \caption{Reasoning instance segmentation~(ReasonSeg-Inst) results among LISA and ours LISA++. All these models use LLaVA v1.5~\cite{liu2023improved} as the base model.}
    \label{table:reason_instseg}   
\vspace{-0.5cm}
\end{table}

\begin{table*}[t!]
    \centering
    \tabcolsep=0.2cm
    {
        \begin{tabular}{ l | c c | c c | c c | c c }
            \toprule
            
            \multirow{3}*{Method} & \multicolumn{2}{c|}{val} & \multicolumn{6}{c}{test} \\ 
            
            \specialrule{0em}{0pt}{1pt}
            \cline{2-9}
            \specialrule{0em}{0pt}{1pt}
            
            
            ~ & \multicolumn{2}{c|}{overall} & \multicolumn{2}{c|}{short query} & \multicolumn{2}{c|}{long query} & \multicolumn{2}{c}{overall} \\

            \specialrule{0em}{0pt}{1pt}
            \cline{2-9}
            \specialrule{0em}{0pt}{1pt}
            
            ~ & gIoU & cIoU & gIoU & cIoU & gIoU & cIoU & gIoU & cIoU \\ 
            
            \specialrule{0em}{0pt}{1pt}
            \hline
            \specialrule{0em}{0pt}{1pt}

            OVSeg~\citep{liang2023open} & 28.5 & 18.6 & 18.0 & 15.5 & 28.7 & 22.5 & 26.1 & 20.8  \\



            GRES~\citep{liu2023gres} & 22.4 & 19.9 & 17.6 & 15.0 & 22.6 & 23.8 & 21.3 & 22.0 \\    %
            
            X-Decoder~\citep{zou2023generalized} & 22.6 & 17.9 & 20.4 & 11.6 & 22.2 & 17.5 & 21.7 & 16.3 \\

            SEEM~\citep{zou2023segment} & 25.5 & 21.2 & 20.1 & 11.5 & 25.6 & 20.8 & 24.3 & 18.7 \\
            
            \specialrule{0em}{0pt}{1pt}
            \hline
            \specialrule{0em}{0pt}{1pt}
            
            LISA-7B & 44.4 & 46.0 & 37.6 & 34.4 & 36.6 & 34.7 & 36.8 & 34.1 \\
            
            LISA-7B (ft) & 52.9 & 54.0 & 40.6 & 40.6 & 49.4 & 51.0 & 47.3 & 48.4 \\
            \specialrule{0em}{0pt}{1pt}
            \hline
            \specialrule{0em}{0pt}{1pt}
            
            LISA-7B-LLaVA1.5 & 53.6 & 52.3 &  47.1  & 48.5 & 49.2 & 48.9  &  48.7 & 48.8 \\
            LISA-7B-LLaVA1.5 (ft) & 61.3 & 62.9 & 48.3  & 46.3  & 57.9  & 59.7 & 55.6 & 56.9\\
            LISA++-7B-LLaVA1.5 (ft) & \textbf{64.2} & \textbf{68.1} & \textbf{49.6} & \textbf{51.1} & \textbf{59.3} & \textbf{61.7} & \textbf{57.0} & \textbf{59.5} \\
            \bottomrule            
        \end{tabular}
    }
    \caption{Reasoning semantic segmentation~(ReasonSeg-Sem) results among LISA, LISA++ (ours) and previous related works. `ft' denotes using 239 reasoning segmentation image-instruction pairs to finetune the model.}
    \label{table:reason_seg}   
\end{table*}

\subsection{Results}
\mypara{Effectiveness on ReasonSeg-Inst.}
We conduct a comparative analysis between LISA++ and the previous version, LISA, on the Reasoning instance segmentation task (ReasonSeg-Inst). The results are presented in Table~\ref{table:reason_instseg}, we utilize AP50, AP75, and mAP as evaluation metrics to assess the performance of our model.

In comparison to LISA-7b, our LISA++-7b achieves significant improvement in ReasonSeg-Inst scenario. Specifically, LISA++ achieves 34.1\%, 22.1\%, and 21.5\% in AP50, AP75, and mAP, respectively. Even when compared to the more powerful LISA-13b, utilizing the 7b LISA++ still resulted in improvements of 18.4\%, 14.3\%, and 12.9\% in AP50, AP75 and mAP, respectively. These results further demonstrate the effectiveness of our model in instance segmentation. We hope LISA++ to serve as the new baseline for the ReasonSeg-Inst task.

\mypara{Effectiveness on ReasonSeg-Sem.}
The ReasonSeg-Sem results are shown in Table~\ref{table:reason_seg}. It is noteworthy that prior studies have not been able to address this task, 
only LISA can accomplish the task involving complex reasoning with more than $20\%$ gIoU performance boost. 
Our newly proposed LISA++, which achieves significant improvement on the ReasonSeg-Inst task, also demonstrates comparable performance to LISA in the original ReasonSeg-Sem task. Specifically, our LISA++ achieves 64.2\% gIoU and 68.1\% cIoU on the validation set, and also achieves 1.4\% gIou and 2.6\% cIoU improvement on the overall testset compared to the LISA, respectively.

These results demonstrate that the model's capability in reasoning instance segmentation does not compromise the original reasoning semantic segmentation. Moreover, it can further improve the model's capacity in ReasonSeg-sem, thereby further validating the generalizability of the LISA Framework.

\begin{figure}[t!]
	\centering
 \includegraphics [width=1\linewidth] 
        {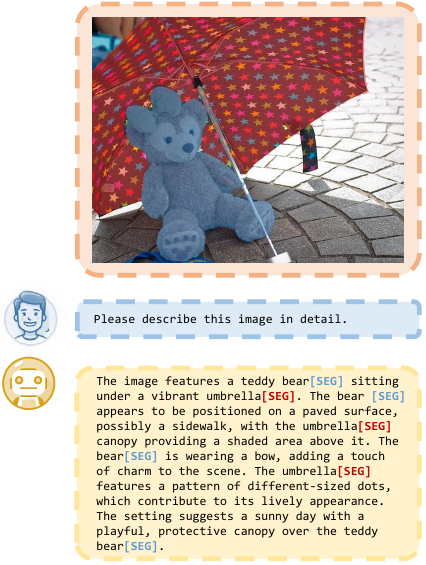}
    
    \caption{Qualitative Results for Image Captioning with the Instance Segmentation.}
    \label{fig:quali_cap}
\end{figure}

\begin{figure*}[t!]
	\centering
        \includegraphics [width=0.9\linewidth] 
        {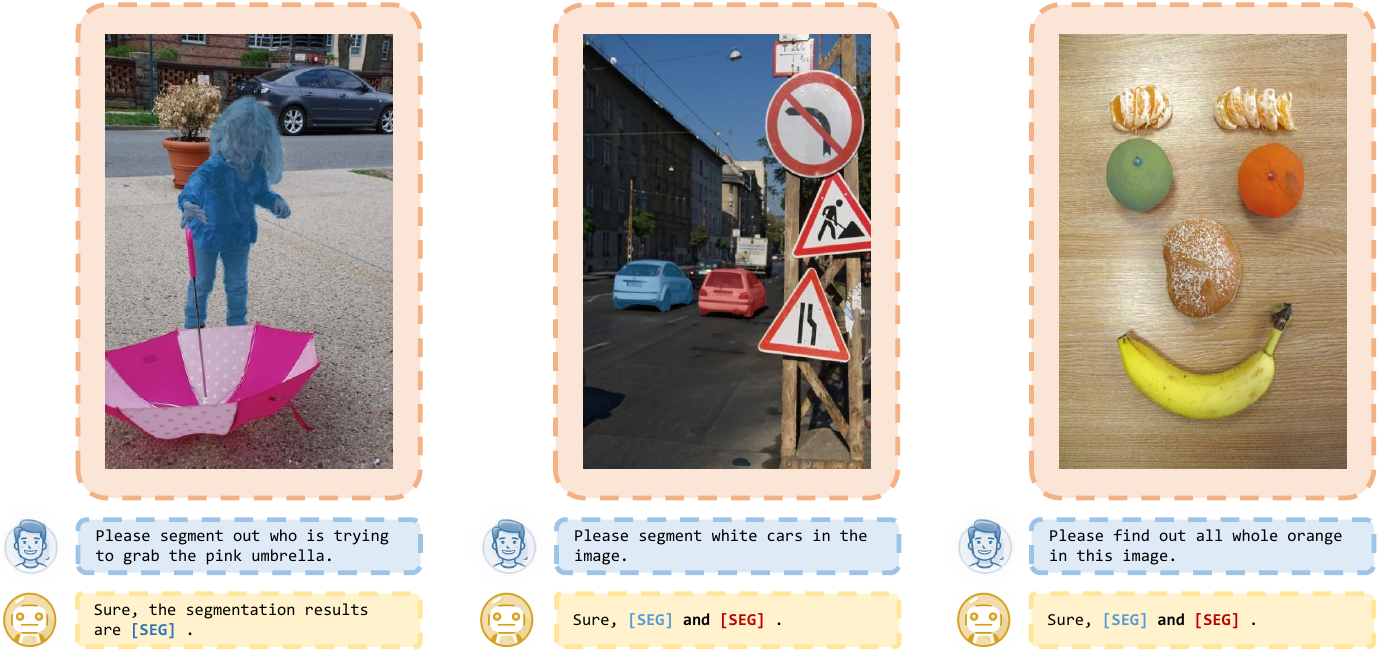}
  
    \caption{Qualitative Results for Instance Segmentation.}
    \label{fig:quali_is}
\end{figure*}
\subsection{Qualitative Results}
To further demonstrate the effectiveness of our LISA++, we show the qualitative results in this section.

\mypara{Image Captioning with Instance Segmentation} As shown in Figure~\ref{fig:quali_cap}, we show LISA++'s qualitative results for Image Captioning with the Instance Segmentation. 
LISA++ can yield more natural and flexible responses, by integrating segmentation results into textual responses, termed the ability of Segmentation during the natural captioning. 

\mypara{Instance Segmentation} Additionally, we assess the instance segmentation performance in Figure~\ref{fig:quali_is}. These questions require advanced reasoning capabilities and pose challenges due to potential interference from other objects. 
Notably, our proposed LISA++ demonstrates exceptional accuracy in recognizing the queried objects and conducting precise segmentation. For instance, it accurately segments objects such as girls, white cars, and whole oranges.

\begin{figure}[t!]
	\centering
        \includegraphics [width=0.85\linewidth] 
        {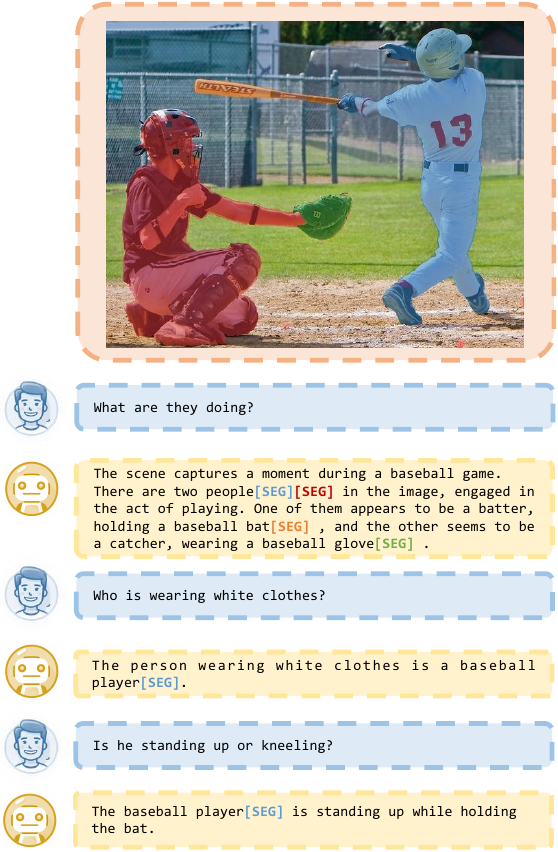}   
 
    \caption{Qualitative Results for Multi-turn Dialogue with the Instance Segmentation.}
    \label{fig:quali_conv}
\end{figure}

\mypara{Multi-turn Dialogue with Instance Segmentation} Furthermore, to conduct a more comprehensive evaluation of LISA++'s performance in the context of multi-turn dialogue with instance segmentation, we employ the "SiD with InstSeg" instructions template and present the qualitative results in Figure~\ref{fig:quali_conv}.
Our proposed LISA++ has the capability to gain a comprehensive understanding of the scene and carry out instance segmentation on the two individuals depicted in the image. Furthermore, LISA++ exhibits well perception of color attributes, such as "white," and spatial relationships, such as "holding the bat."

\section{Conclusion}
\label{sec:conclusion}
In this work, we present LISA++ by incorporating the reasoning instance segmentation and segmentation dialogue abilities into LISA, enabling more natural and flexible conversations with segmentation outcomes. This study demonstrates that the mask-as-embedding paradigm, introduced in the original LISA, can be effectively scaled to accommodate more complex scenarios without structural modifications. We hope that this paradigm, with the advancements in LISA++, can offer new perspectives and stimulate further innovative research within the academic community.

\begin{table*}[t]
\caption{The prompt for yielding instruction-tuning data for Reasoning Instance Segmentation (ReasonSeg-Inst).}
\label{tab:prompt_reasonseg_inst}
\centering
\begin{minipage}{1.5\columnwidth}\vspace{0mm}    \centering
\begin{tcolorbox} 
    \centering
    \small
     \hspace{-6mm}
    \begin{tabular}{p{0.99\columnwidth}}

\begin{minipage}{0.99\columnwidth}\vspace{0mm}
\VarSty{messages} = [\{\var{"role": "system", "content":} 
You are asked to generate the instruction tuning data for language-guided reasoning instance segmentation. Requirements are: 
(1) Create a series of specific questions (Q1, Q2, Q3, etc.)(but no more than 5 questions) focusing on identifying and isolating different elements within the image, based on the polygon information. Each question should not refer to previous questions, and facilitate the generation of segmented masks for objects when processed by an imaging system. Ensure the questions are clear, precise, logical, and interesting, and avoid directly mentioning coordinates, label names, and polygons. The questions should try to consider the use and nature of the object, not just its appearance. The output format must be `Q[number]: [question]'. If the question is about humans, do not ask questions without extra modifiers, but ask questions simply like `Please find out all the individuals in the image.'
(2) Answer all your questions (A1, A2, A3, etc.) indicating which polygons in ~\textless~anno~\textgreater correspond to each question. For items with multiple instances in the same category, list ALL instances for that category in the answer! Do not output full information; the format MUST follow: `A[number]: instance id is [id1], label name is [name]; instance id is [id2], label name is [name]; instance id is [id3], label name is [name]; ...'
 \}
]
        
\end{minipage}
    \end{tabular}
\end{tcolorbox}

\vspace{-2mm}
\end{minipage}
\end{table*}

\begin{table*}[t]
\caption{The prompt for yielding instruction-tuning data for Q\&A with Segmentation in Dialogue (SiD).}
\label{tab:prompt_qa}
\centering
\begin{minipage}{1.5\columnwidth}\vspace{0mm}    \centering
\begin{tcolorbox} 
    \centering
    \small
     \hspace{-6mm}
    \begin{tabular}{p{0.99\columnwidth}}

\begin{minipage}{0.99\columnwidth}\vspace{0mm}
\VarSty{messages} = [\{\var{"role": "system", "content":} 
You are asked to generate the Q\&A conversational data. Requirements are: 
(1) Construct a dialogue that paints a vivid picture of the scene through natural and diverse questions and answers, ensuring a logical and engaging flow. 
(2) The context of the dialogue can be relevant. Include interactions that cover object identification, counting, actions, locations, and the relationship between objects, while also integrating complex queries that delve into the objects' background information and the scenario depicted in the image. 
(3) Carefully formulate questions to avoid ambiguity and ensure they can be answered with confidence based on the image annotations. Avoid including ~\textless~instance id; label name~\textgreater in ~\textless~person~\textgreater's queries. Do not directly mention `polygon', or `annotations' in the questions and answers. 
(4) Format the output as:
`~\textless~person~\textgreater: XXXX 
~\textless~robot~\textgreater: XXXX'
, with ~\textless~robot~\textgreater responses incorporating instance IDs and label names like `keyboards ~\textless~34494; keyboard~\textgreater ~\textless~31264; keyboard~\textgreater'. 
 \}
]
        
\end{minipage}
    \end{tabular}
\end{tcolorbox}

\vspace{-2mm}
\end{minipage}
\end{table*}

\begin{table*}[t]
\caption{The prompt for yielding instruction-tuning data for captioning with Segmentation in Dialogue (SiD).}
\label{tab:prompt_caption}
\centering
\begin{minipage}{1.5\columnwidth}\vspace{0mm}    \centering
\begin{tcolorbox} 
    \centering
    \small
     \hspace{-6mm}
    \begin{tabular}{p{0.99\columnwidth}}

\begin{minipage}{0.99\columnwidth}\vspace{0mm}
\VarSty{messages} = [\{\var{"role":"system", "content":} 
You are asked to generate the captioning conversational data.
Please generate one question-and-answer pair based on the provided image (image\_size: \{image\_size\}) and its instance segmentation annotation. The focus is on describing(captioning) the whole image focusing on those instances given in the annotation, as detailedly as you can without directly referencing anything in the annotation except for instance id. Make sure the answers indicate the specific instances involved. The annotation consists of struct \{`label name', `instance id', `bbox', `center point'\} that each is corresponded with a unique instance in the image (segmentation mask is given in the form of bbox[left, top, right, bottom] and `center\_point' is the center of the instance. x-coordinates are increasing from left to right. y-coordinates are increasing from top to bottom! The more the instance is close to the TOP edge, the SMALLER the y-coordinate is.
Please assume that you are in a space where point1~[0, 0] is to the upper left of point2~[1, 1], and point2~[1, 1] is to the bottom right of point1~[0, 0]. The starting point~[0, 0] is on the top-left of the given image. The generated QA should follow the format of:
`Q1: ~\textless~question~\textgreater.
A1: ~\textless~descriptions~\textgreater' \}
]
        
\end{minipage}
    \end{tabular}
\end{tcolorbox}

\vspace{-2mm}
\end{minipage}
\end{table*}

\clearpage
{
    \small
    \bibliographystyle{ieeenat_fullname}
    \bibliography{mybib}
}


\end{document}